\documentclass[conference]{IEEEtran}
\IEEEoverridecommandlockouts
\usepackage{amsmath,amsfonts}
\usepackage{algorithmic}
\usepackage[ruled]{algorithm2e}
\usepackage{subfigure}
\usepackage{array}
\usepackage{textcomp}
\usepackage{stfloats}
\usepackage{url}
\usepackage{verbatim}
\usepackage{graphicx}
\usepackage{cite}
\usepackage{booktabs}
\usepackage{makecell}
\usepackage{multirow}
\usepackage{balance}
\usepackage{amssymb}
\usepackage{bbding}
\usepackage{pifont}
\def\BibTeX{{\rm B\kern-.05em{\sc i\kern-.025em b}\kern-.08em
    T\kern-.1667em\lower.7ex\hbox{E}\kern-.125emX}}
\usepackage{hyperref}
\hypersetup{
    colorlinks=true,
    linkcolor=blue,
    anchorcolor=blue,
    citecolor=blue}
\usepackage{orcidlink}
\begin{document}

\title{TRGR: Transmissive RIS-aided Gait Recognition Through Walls\\
}

\author{\IEEEauthorblockN{Yunlong Huang\textsuperscript{1}, Junshuo Liu\textsuperscript{1}, Jianan Zhang\textsuperscript{1}, Tiebin Mi\textsuperscript{1}, Xin Shi\textsuperscript{2}, Robert Caiming Qiu\textsuperscript{1}}
\IEEEauthorblockA{\textsuperscript{1} School of Electronic Information and Communications, Huazhong University of Science and Technology, Wuhan, China \\
\textsuperscript{2} School of Control and Computer Engineering, North China Electric Power University, Beijing, China \\
Email:\{huangyunlong, junshuo\_liu, zhangjn, mitiebin, caiming\}@hust.edu.cn, xinshi@ncepu.edu.cn} 
}

\maketitle

\begin{abstract}
Gait recognition with radio frequency (RF) signals enables many potential applications requiring accurate identification. 
However, current systems require individuals to be within a line-of-sight (LOS) environment and struggle with low signal-to-noise ratio (SNR) when signals traverse concrete and thick walls. 
To address these challenges, we present TRGR, a novel transmissive reconfigurable intelligent surface (RIS)-aided gait recognition system. TRGR can recognize human identities through walls using only the magnitude measurements of channel state information (CSI) from a pair of transceivers.
Specifically, by leveraging transmissive RIS alongside a configuration alternating optimization algorithm, TRGR enhances wall penetration and signal quality, enabling accurate gait recognition. Furthermore, a residual convolution network (RCNN) is proposed as the backbone network to learn robust human information.
Experimental results confirm the efficacy of transmissive RIS, highlighting the significant potential of transmissive RIS in enhancing RF-based gait recognition systems.
Extensive experiment results show that TRGR achieves an average accuracy of 97.88\% in identifying persons when signals traverse concrete walls, demonstrating the effectiveness and robustness of TRGR.
\end{abstract}

\begin{IEEEkeywords}
Gait analysis, RF sensing, through-wall sensing, transmissive RIS, neural network
\end{IEEEkeywords}

\section{Introduction}
Gait recognition has played an important role within the realm of intelligent Internet of Things (IoT) applications and smart environments. Device-free RF sensing in gait recognition has been an emerging research trend in the past years, which recognizes individuals from a distance without requiring them to perform any specific active task (e.g., fingerprint scanning\cite{ali2016overview}). Furthermore, compared to other methods, RF-based gait recognition is more promising, which does not require a direct visual line of sight of the walking person and is not affected by lighting conditions. 

Early studies in RF-based gait recognition relied on specialized radar technology and broad-spectrum signals to analyze the way people walk. For instance, Mehmood \textit{et al.} \cite{Mehmood2010} used radar to measure stride rates and body segment velocities, while Wang \textit{et al.} \cite{Wang2014} applied Doppler radar to capture walking speed and step duration. Additionally, Yardibi \textit{et al.} \cite{5766971} developed algorithms for extracting gait features from pulse-Doppler radar spectrograms, enabling the estimation of gait velocity and stride lengths. 

Recently, leveraging CSI for gait recognition has drawn enormous research efforts, showcasing its potential in identifying individuals. WiFiU \cite{wang2016gait} utilized Wi-Fi CSI to create spectrogram-based gait features for individual classification. WiWho \cite{7460727} explored step-and-walk analysis through CSI for gait identification, while WiFi-ID \cite{7536315} analyzed CSI perturbations to discover unique identification features. These studies highlight the growing importance of CSI as a non-invasive tool for biometric identification in gait recognition research.

Existing gait recognition systems face challenges with RF signal attenuation through concrete and thick walls, leading to low SNR and impairing recognition accuracy. To overcome these challenges, we propose the use of transmissive reconfigurable intelligent surfaces \cite{tang2023transmissive}. This solution allows for the precise adjustment of signal phase and amplitude through reconfigurable meta-atoms, converting unfavorable propagation paths into optimal ones. This adaptive control mechanism significantly enhances transmission quality, facilitating more effective passive gait recognition.

In this paper, we present a transmissive RIS-aided gait recognition system, a pioneering solution capable of recognizing human identities in challenging environments characterized by concrete structures and thick walls.
The main contributions of this paper can be summarized as follows:
\begin{itemize} 
    \item \textbf{We present a transmissive RIS-assisted through-wall gait recognition system, leveraging the transmissive RIS to enhance RF signal penetration through walls, thereby enabling accurate gait recognition. To the best of our knowledge, this is the first work to leverage the transmissive RIS to enable gait recognition through walls.}

    \item \textbf{We design a residual convolutional neural network (RCNN) specifically designed for gait recognition in our TRGR system, showcasing its effectiveness as the backbone network for gait recognition.}
    
    \item \textbf{We implement a prototype of TRGR on a pair of transceivers and conduct experiments in real-world scenarios. The experimental results demonstrate a notable accuracy of 97.88\%, indicating the effectiveness and suitability of TRGR for continuous and widespread application.}
\end{itemize}
\section{Preliminary}
\subsection{Channel State Information}
Channel state information is crucial for characterizing Wi-Fi link attributes at the physical layer, particularly in reflecting the channel frequency response to multipath fading and shading.
In the indoor environment, human movement significantly influences CSI, revealing detailed effects of delay, amplitude decay, and phase shift across communication subcarriers\cite{yang2018device}. The frequency domain of the RF signal is modeled as the channel impulse response $h(\tau)$: 
\begin{equation}
\begin{aligned}
\\h(\tau) &= \overbrace{\sum_{m=1}^M\alpha_me^{j\phi_m}\delta(\tau-\tau_m)\quad}^{\text{from Tx to Rx without utilizing RIS}} \\
 &+ \overbrace{\sum_{n=1}^N\alpha_n\beta_{n}e^{j(\phi_n+\theta_n)}\delta(\tau-\tau_n)\quad}^{\text{from Tx to Rx utilizing RIS}}
\end{aligned}
\end{equation}
where $M$ and $N$ represent the multipath components from the transmitter (Tx) to the receiver (Rx) without and with transmissive RIS, respectively, $\alpha_m$ and $\phi_m$ represent the amplitude and phase of the $m$th multipath component from Tx to Rx without utilizing transmissive RIS, respectively, $\delta(\tau)$ denotes the Dirac delta function, and $\tau_m$ and $\tau_n$ denote the time delay, $\alpha_n$ and $\phi_n$ represents the amplitude and phase of the $n$th multipath component from Tx to Rx through transmissive RIS, $\beta_{n}=1$ denotes the amplitude shift of the $n$th multipath component from Tx to Rx through transmissive RIS, $\theta_n$ denotes the phase shift of the $n$th multipath component from Tx to Rx through transmissive RIS.

In real-world scenarios, the signal strength from the transmitter to the receiver without RIS through walls is typically weak and susceptible to noise interference. Consequently, our focus here is primarily on analyzing the signal transmission from the transmitter to the receiver through transmissive RIS, highlighting the role of transmissive RIS in enhancing sensing in obstructed environments.

\subsection{Transmissive Reconfigurable Intelligent Surfaces}
\begin{figure}[!ht]
\centerline{\includegraphics[width=1\columnwidth]{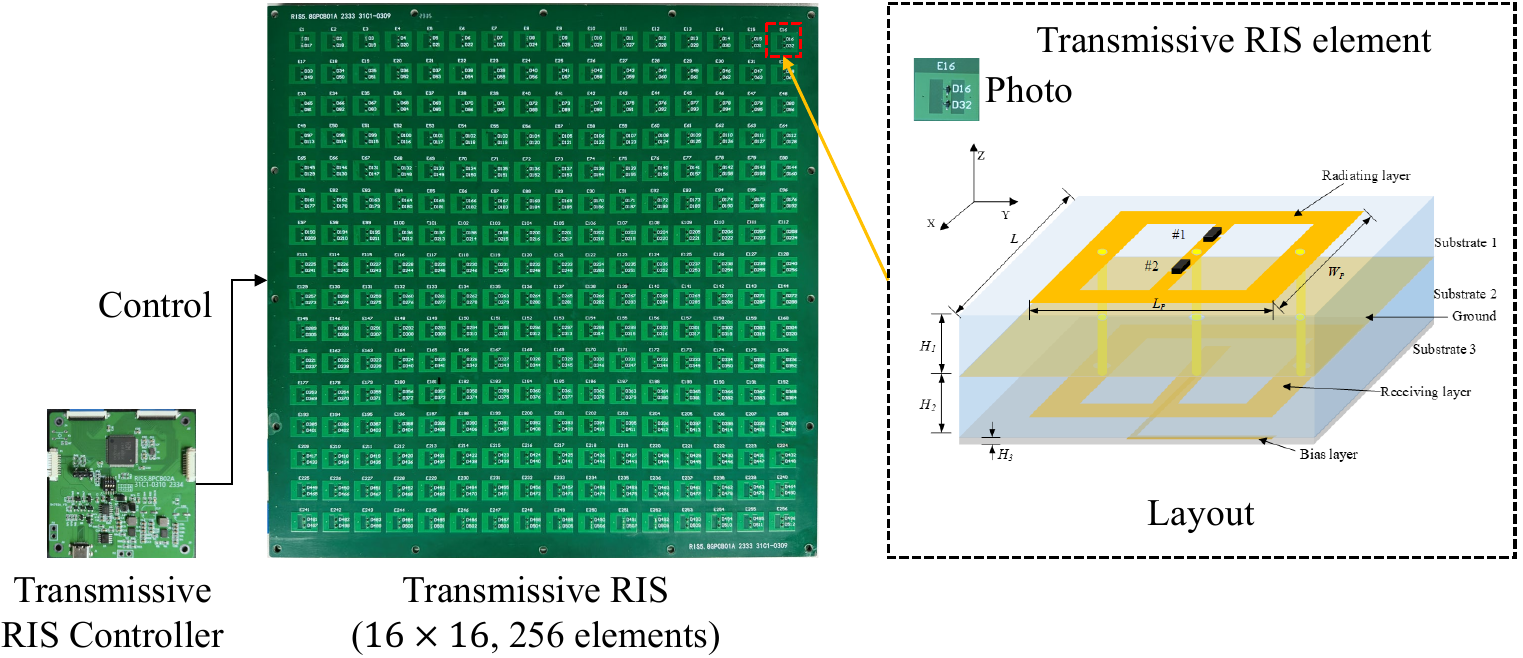}}
\caption{Logic circuit controlling board (left side), photography of the 1-bit transmissive RIS prototype (middle side), and the layout of the transmissive RIS element (right side).}
\label{figure:tris}
\end{figure}

We utilize an electrically modulated transmissive RIS, as depicted in Fig.~\ref{figure:tris}. 

The transmissive RIS utilized in this study measures $28.8\times28.8\times0.3\;\mathrm{cm}^{3}$ and comprises a two-dimensional array of electrically controllable RIS elements. This array is structured such that each row and column contains 16 RIS elements, culminating in a total of 256 elements.

Each RIS element, measuring $1.8\times1.8\times0.3\;\mathrm{cm}^{3}$, comprises four copper layers supported by three substrate layers. It includes two PIN diodes for state modulation: in the 0-state, PIN diode \#1 is off and PIN diode \#2 is on; for the 1-state, this configuration is reversed, with PIN diode \#1 on and PIN diode \#2 off. This binary setup enables dynamic control over the electromagnetic behavior of RIS for precise wave manipulation.

\begin{algorithm}[!ht]
	\caption{Configuration Optimization Algorithm\cite{pei2021ris}}  
	\label{alg:config_opt}
	\KwIn{The current signal strength $\mathbf{s}_{current}$}
	\KwOut{$\mathbf{\Phi}$}  
    \textbf{Initialization:} Initialize variables $\mathbf{\Phi}$, $\mathbf{s}_{max}$, $col$, $row$, $\mathbf{\Phi}_{current}$.\\
	\For{t=1 to 5}{
 	\For{i=1 to 32}{
            \lIf {$i\le16$}{$col=col+1$}       
            \lElse {$row=row+1$}
            Obtain the current signal strength $\mathbf{s}_{current}$.\newline
            \lIf {$\mathbf{s}_{current}>\mathbf{s}_{max}$}{$\mathbf{s}_{max}=\mathbf{s}_{current},\mathbf{\Phi}=\mathbf{\Phi}_{current}$}       
            Utilize the greedy algorithm to obtain the subsequent optimized codebook $\mathbf{\Phi}_{next}$, $\mathbf{\Phi}_{current}=\mathbf{\Phi}_{next}$.\newline
            Assign the codebook $\mathbf{\Phi}_{current}$ to the transmissive RIS.\newline
	   }
	}
\end{algorithm} 

To elucidate the optimization process of the RIS configuration, the model for the received signal can be elaborated as follows:
\begin{equation}
y=\mathbf{h}\mathbf{\Phi}\mathbf{H}x+w
\label{eq:model}
\end{equation}
where $x$ symbolizes the transmitted signal, $\mathbf{\Phi}$ represents the phase shift matrix implemented at the transmissive RIS, and $w$ denotes the zero-mean additive white Gaussian noise (AWGN) with a variance of $\sigma^2$. The matrices $\mathbf{H}$ and $\mathbf{h}$ respectively signify the channel matrices from the transmitter to the transmissive RIS, and from the transmissive RIS to the receiver. The SNR is defined by
\begin{equation}
\rho=\frac{|\mathbf{h}\mathbf{\Phi}\mathbf{H}|^2}{\sigma^2}
\end{equation}
The optimization problem can be formulated as
\begin{equation}
 \begin{aligned}
 \max_{\mathbf{\Phi}} \quad & \rho(\mathbf{\Phi}) \\
\text {subject to} \quad&|\phi_n|=1, \forall n=1,2,\ldots,N. \\ 
 \end{aligned}
\end{equation}

Utilizing a greedy algorithm\cite{pei2021ris}, we iteratively obtain the local optimal codebook configuration, progressively enhancing the transmissive RIS configuration. This process, named the configuration alternating optimization algorithm (detailed in Algorithm~\ref{alg:config_opt}), systematically improves the signal handling capability of RIS.

\section{System Design}
\begin{figure*}[!ht]
\centering
\includegraphics[width=1\linewidth]{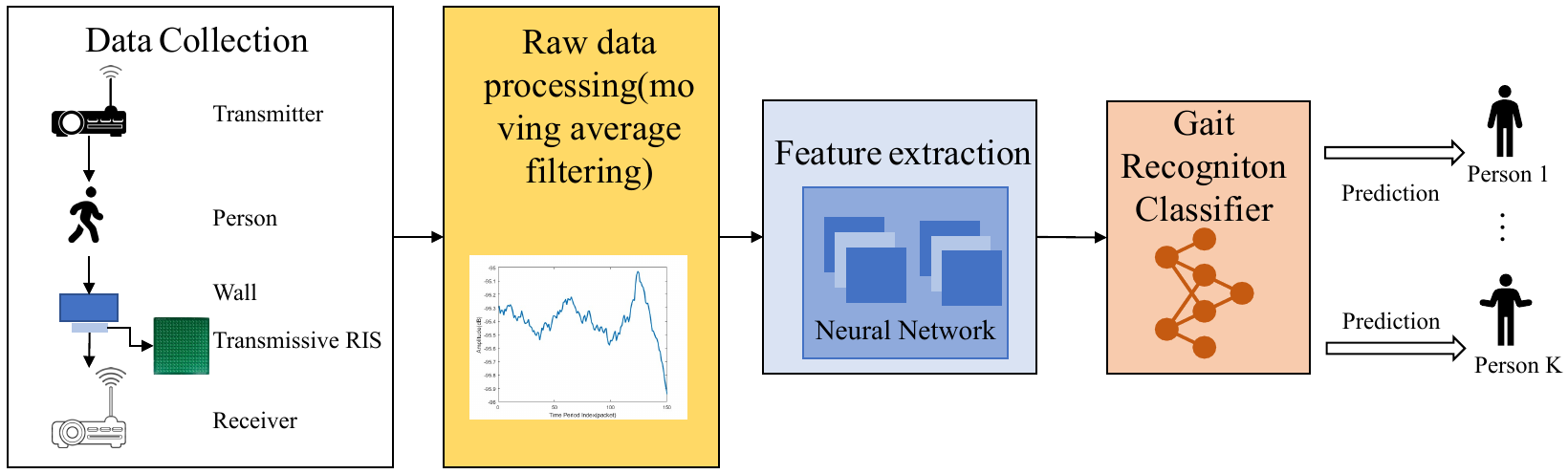}
\caption{Architecture of the proposed transmissive RIS-aided through-wall gait recognition system.}
\label{figure:system_overview}
\end{figure*}

In this paper, we propose a novel system TRGR for gait recognition through walls using a transmissive RIS. Our system aims to identify individuals based on the magnitude of CSI measurements obtained from a pair of transceivers. The system architecture, as illustrated in Fig.~\ref{figure:system_overview}, comprises four main modules: data collection, raw data processing, feature extraction, and gait recognition. 

The data collection module of TRGR creates a CSI monitoring environment with a transmitter and a receiver to gather CSI data on gait patterns from various volunteers. The raw CSI data is then processed in the raw data processing module to enhance noise removal and feature extraction. Subsequently, in the feature extraction and gait recognition module, a neural network is trained by TRGR to identify individuals based on the CSI data associated with different volunteers.

\subsection{System Hardware Description}
\begin{figure}[!ht]
\centerline{\includegraphics[width=1\columnwidth]{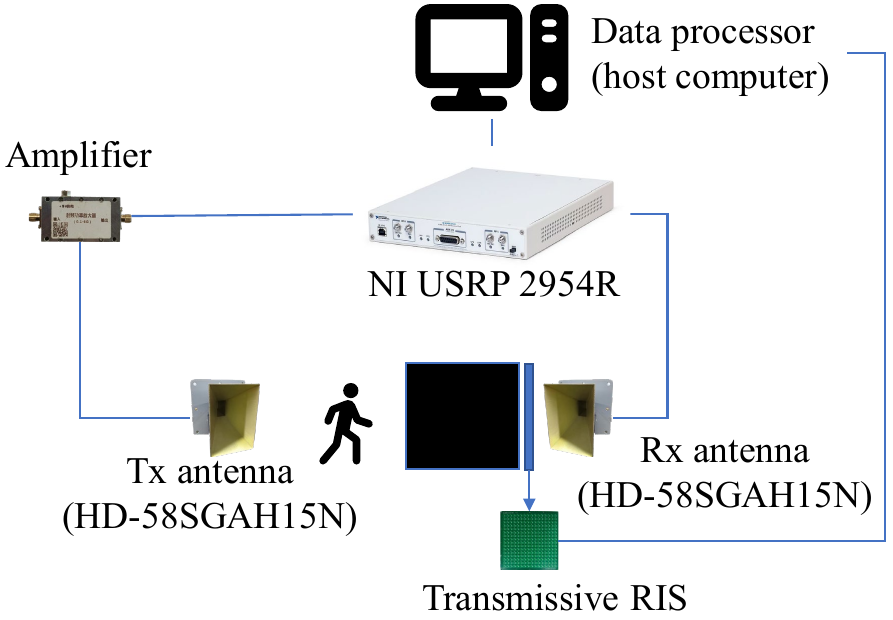}}
\caption{ Components of the transceiver module.}
\label{figure:transceiver_module}
\end{figure}

The system operates at the 5.8GHz commodity Wi-Fi frequency with a 160MHz bandwidth using continuous wave signal. It includes signal transceiver equipment Universal Software Radio Peripheral (USRP)\cite{ettus2015universal}, a well-designed transmissive RIS, a transmitter antenna, and a receiver antenna. The USRP is equipped with one transmitter and one receiver for signal transmission and reception. The transmissive RIS serves to aid in wall penetration and boost the signal-to-noise ratio at the receiver. The transmissive RIS comprises $16\times16$ passive antenna units and the status of each unit can be controlled via on/off diodes. The host computer can control transmissive RIS by transmitting the status of each RIS unit to the microcontroller for control.
As shown in Fig.~\ref{figure:transceiver_module}, the transceiver module of the system consists of the following components:
\begin{itemize}  
    \item \textbf{Tx-Rx USRP device:} The transmitter and receiver are developed using a USRP (NI USRP-2954R), capable of converting baseband to RF signals and vice versa. Equipped with RF modulation/demodulation circuits and a baseband processing unit, the USRP allows for software-based management, providing flexibility in signal processing \cite{ettus2015universal}.
    \item \textbf{Tx and Rx antennas:} The system utilizes directional horn antennas (HD-58SGAH15N) for both Tx and Rx, with both antennas featuring linear polarization.
    \item \textbf{Amplifier:} To cope with the large attenuation incurred by the reflections on the wall, we connect one amplifier to the Tx antenna, which amplifies the transmitted RF signals of the USRP.
    \item \textbf{Data processor:} The host computer, utilizing LabVIEW \cite{Kodosky2020}, controls the USRP and extracts measurement vectors from its received signals. These vectors are then analyzed by a neural network to identify individuals.
\end{itemize}

\subsection{Denoising}
\begin{figure}[!ht]
\centering
\subfigure[The waveform of the subcarrier before denoising.]
{
    \begin{minipage}[b]{0.45\linewidth}
        \centering
        \includegraphics[width=1\linewidth]{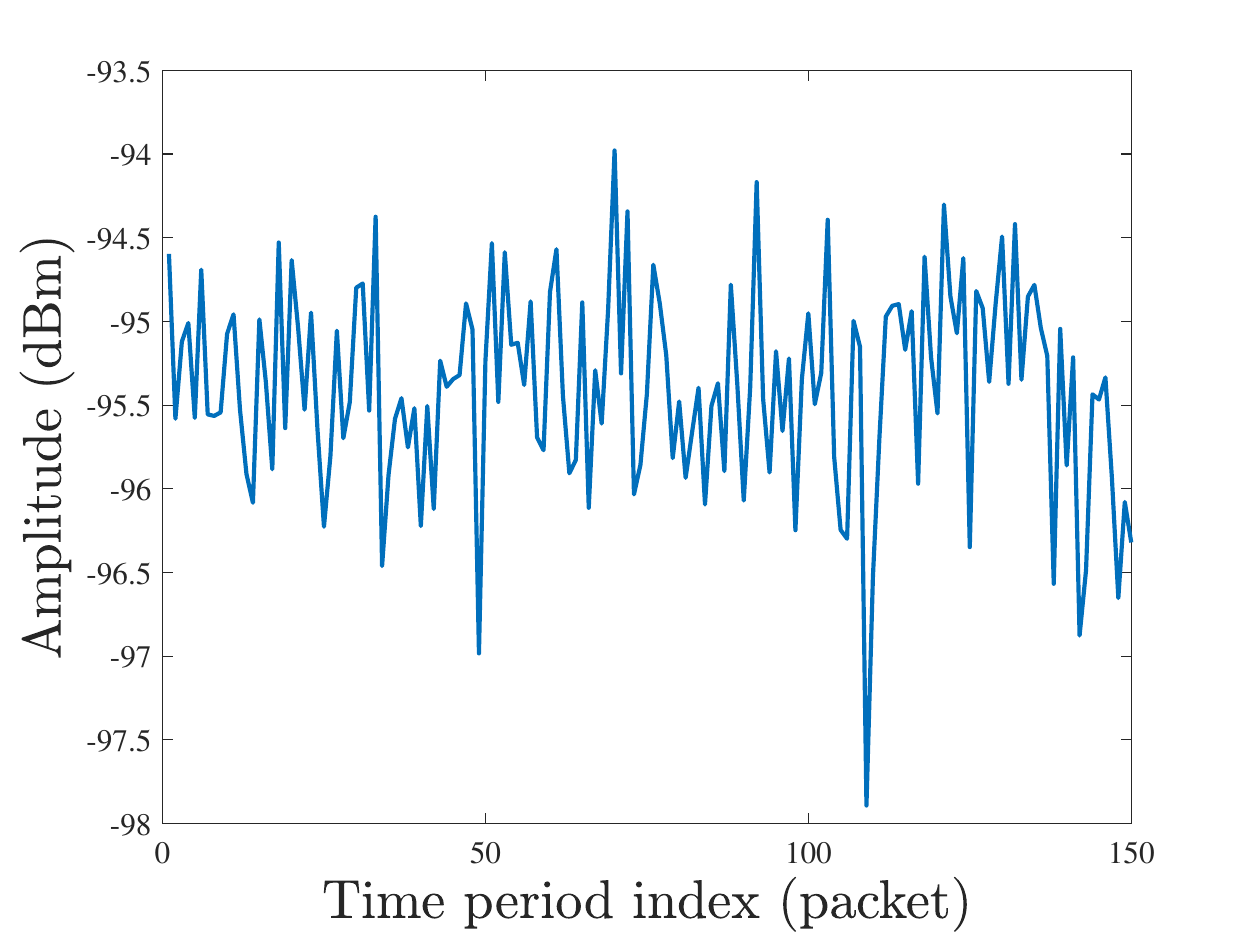}
    \end{minipage}
}
\subfigure[The waveform of the subcarrier after denoising.]
{
    \begin{minipage}[b]{0.45\linewidth}
        \centering
        \includegraphics[width=1\linewidth]{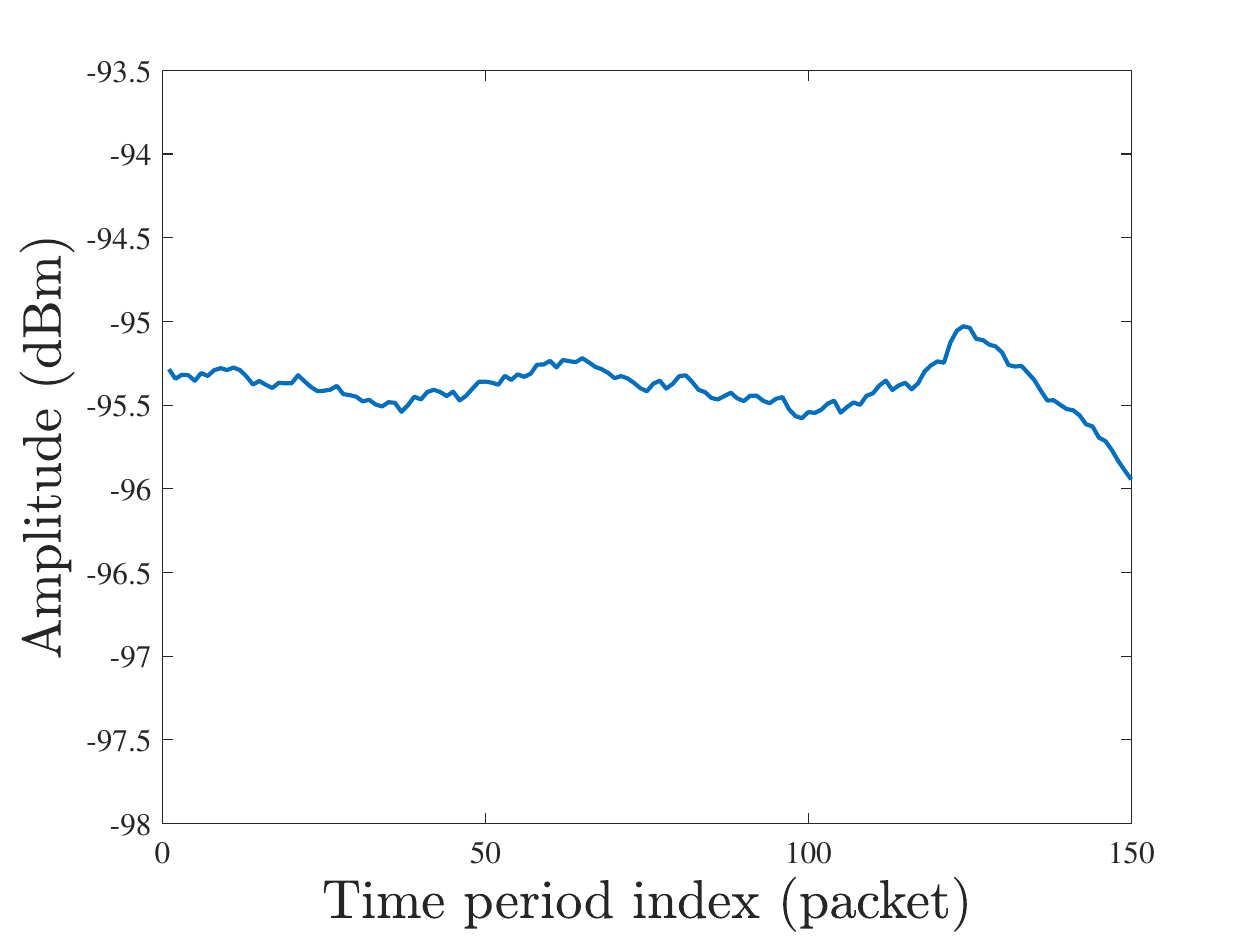}
    \end{minipage}
}
\caption{Comparison of raw and denoising data on the 5200th subcarrier in
the CSI stream.}
\label{figure:movmean_compare}
\end{figure}

In the denoising module, we choose the common denoising method of moving average filter method \cite{7581108} to eliminate noise. 
In order to analyze the effect of the moving average filter method on CSI, we visualize one of the CSI subcarriers (the 5200th out of 8192) for analysis. Fig.~\ref{figure:movmean_compare} compares the waveforms of the raw data and the data after the moving average filter method, where the $y-axis$ is the amplitude of the 5200th subcarrier and the $x-axis$ is the index of the time packet. The figure demonstrates that moving average filtering can effectively eliminate noise.

\subsection{Data Analysis}
\begin{figure*}[!ht]
    \centering
    {Subject 1}
	\begin{minipage}{0.22\linewidth}Sample 1
		\centering
		\includegraphics[width=1\linewidth]{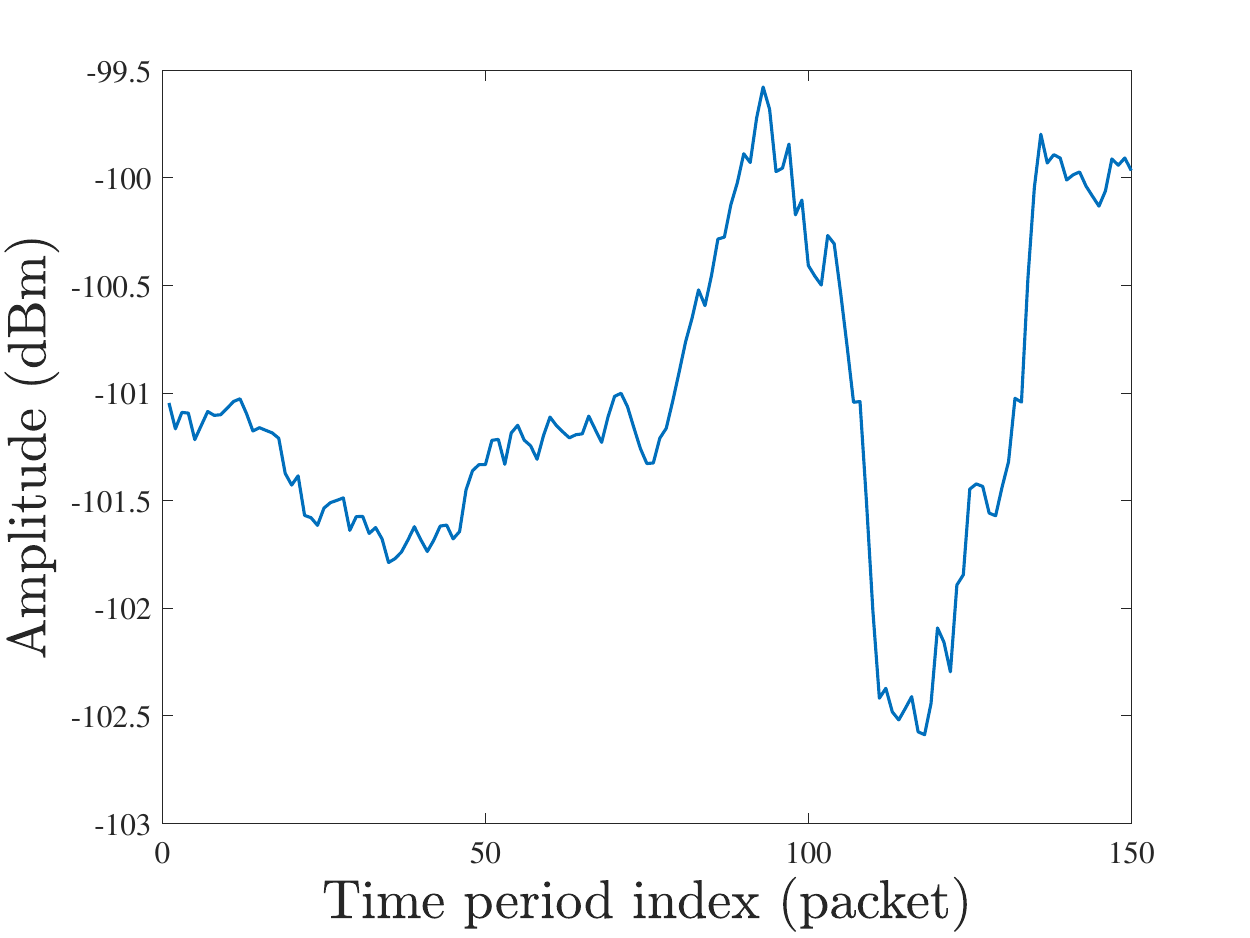}
	\end{minipage}
	\begin{minipage}{0.22\linewidth}Sample 2
		\centering
		\includegraphics[width=1\linewidth]{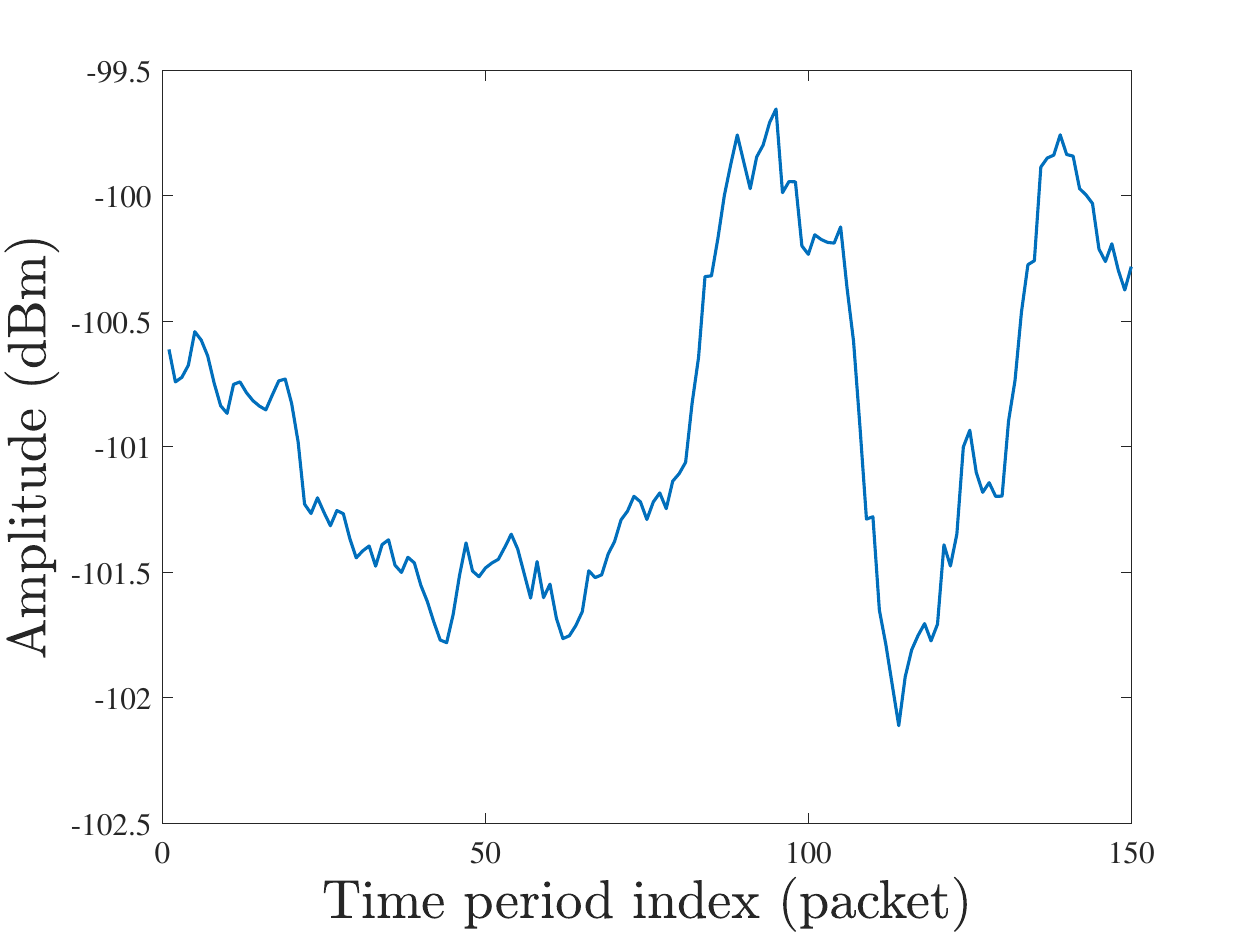}
	\end{minipage}
	\begin{minipage}{0.22\linewidth}Sample 1 \& 2
		\centering
		\includegraphics[width=1\linewidth]{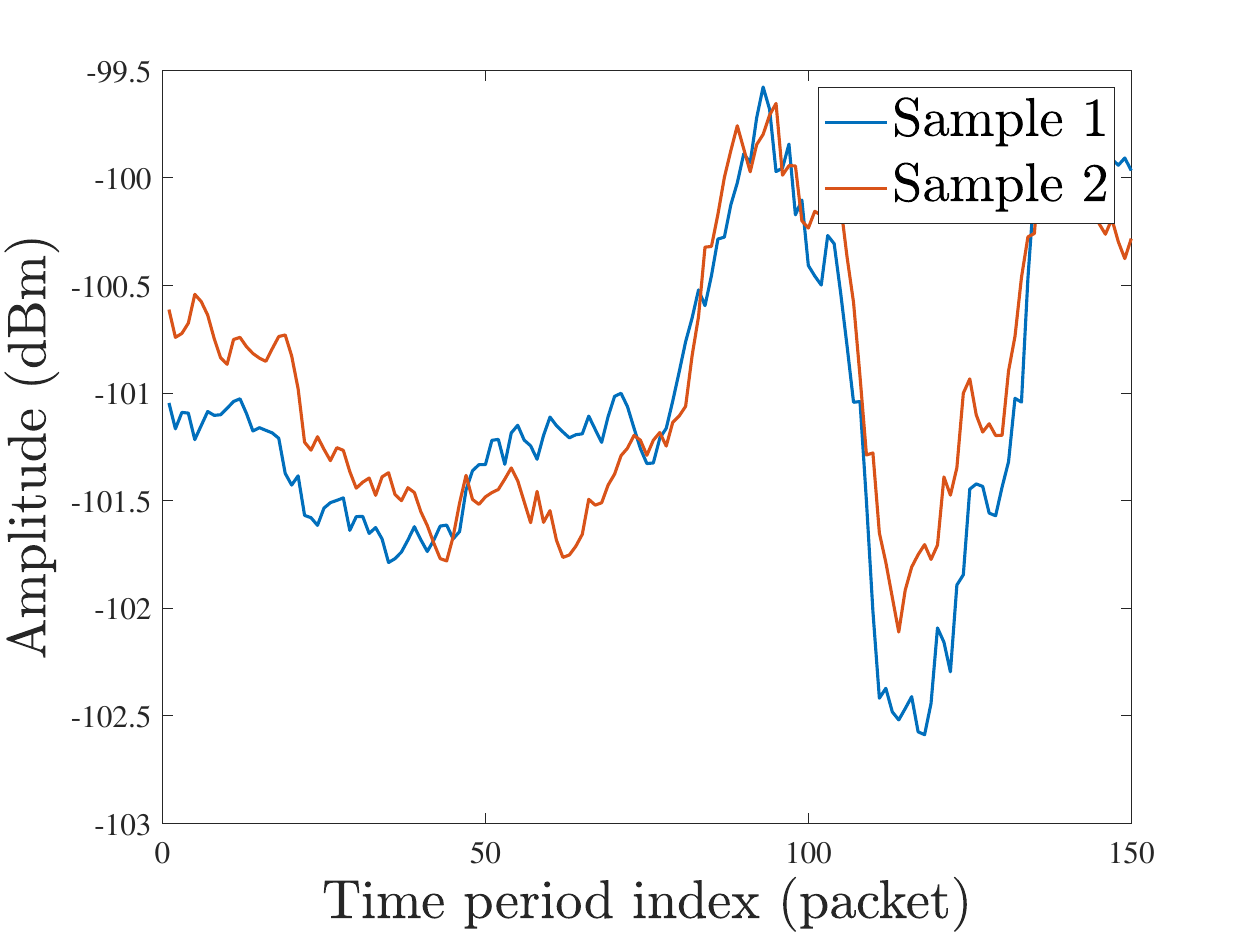}
	\end{minipage}
	\begin{minipage}{0.22\linewidth}Sample 1 \& No subject
		\centering
		\includegraphics[width=1\linewidth]{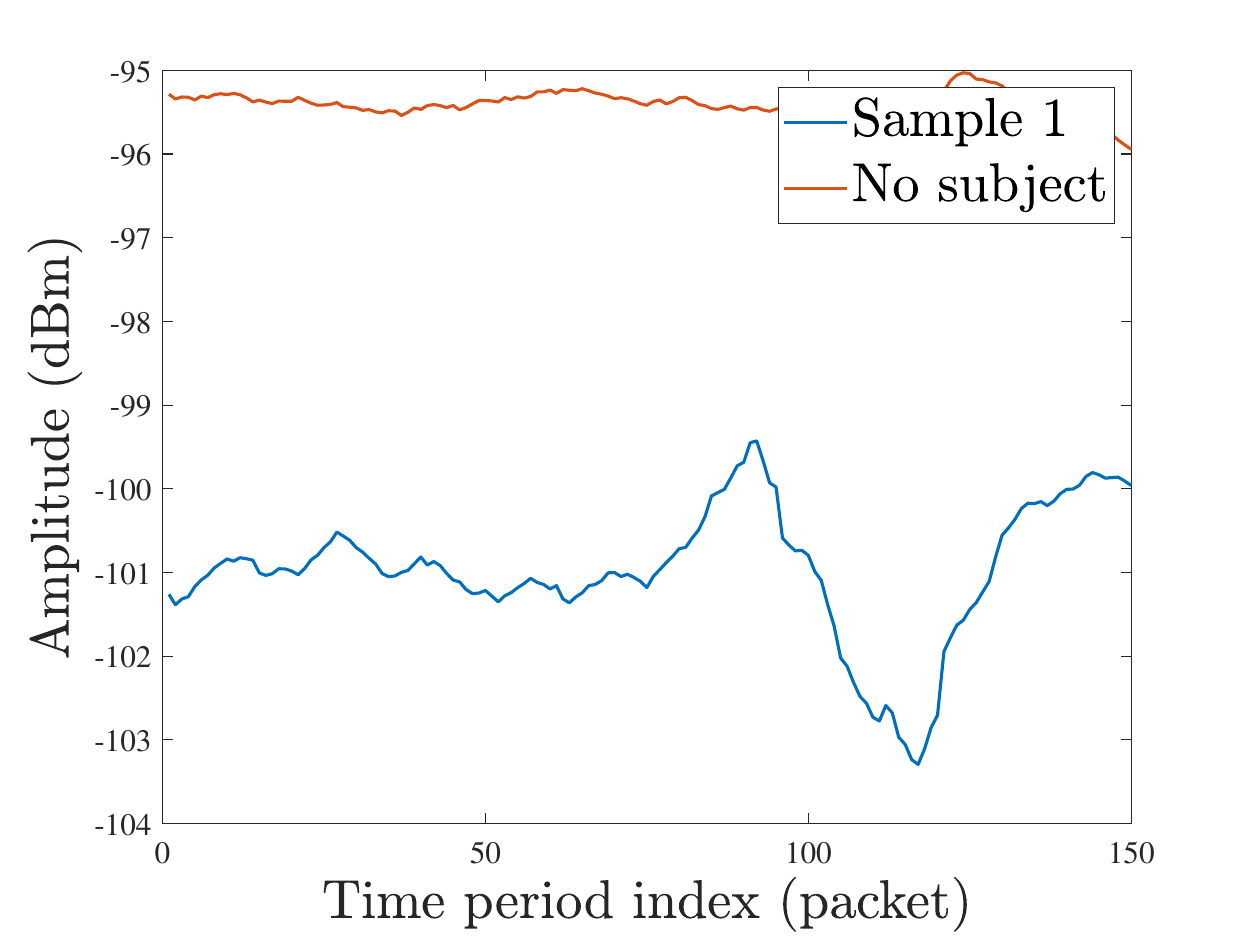}
	\end{minipage}
    \qquad
    {Subject 2}
	\begin{minipage}{0.22\linewidth}
		\centering
		\includegraphics[width=1\linewidth]{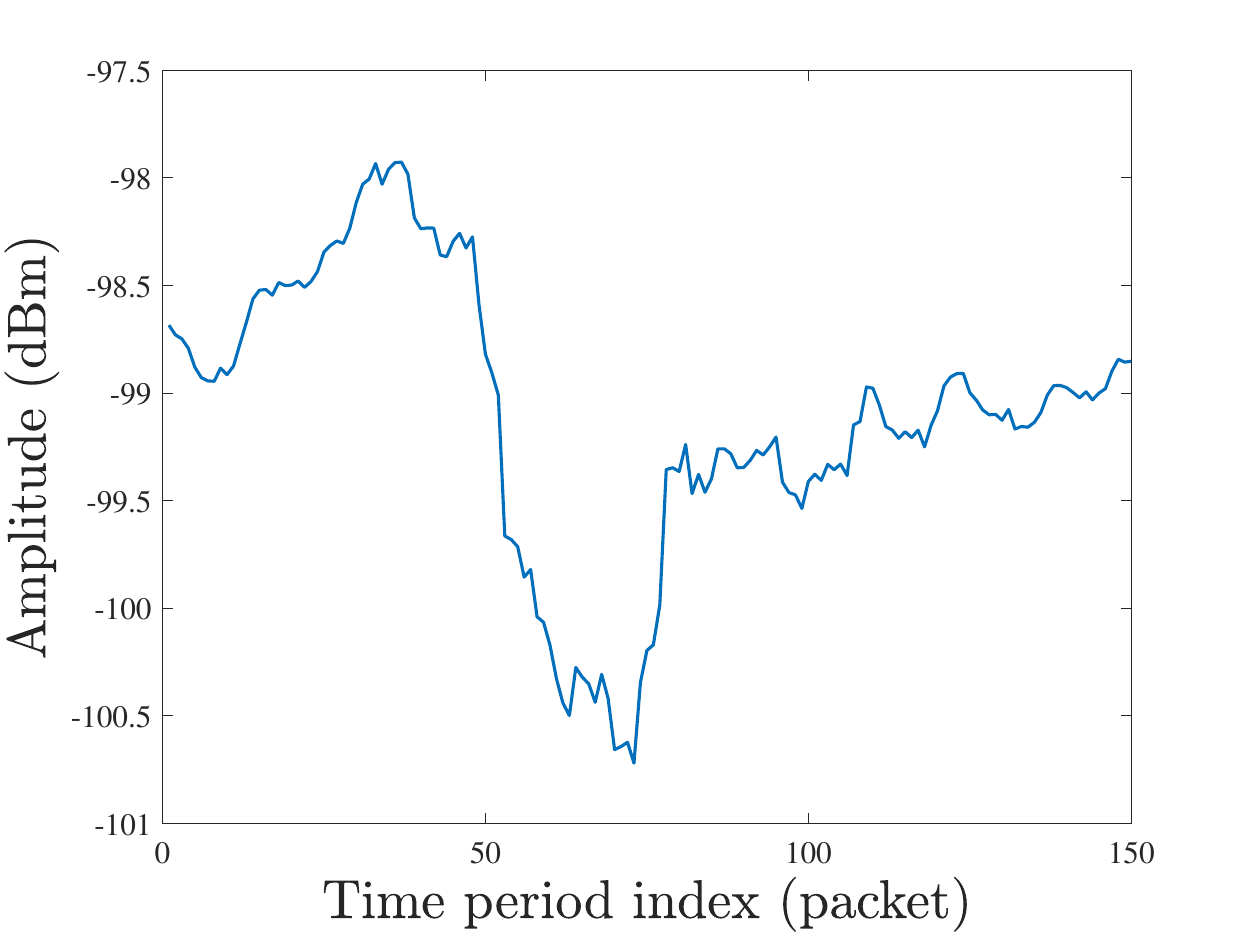}
	\end{minipage}
	\begin{minipage}{0.22\linewidth}
		\centering
		\includegraphics[width=1\linewidth]{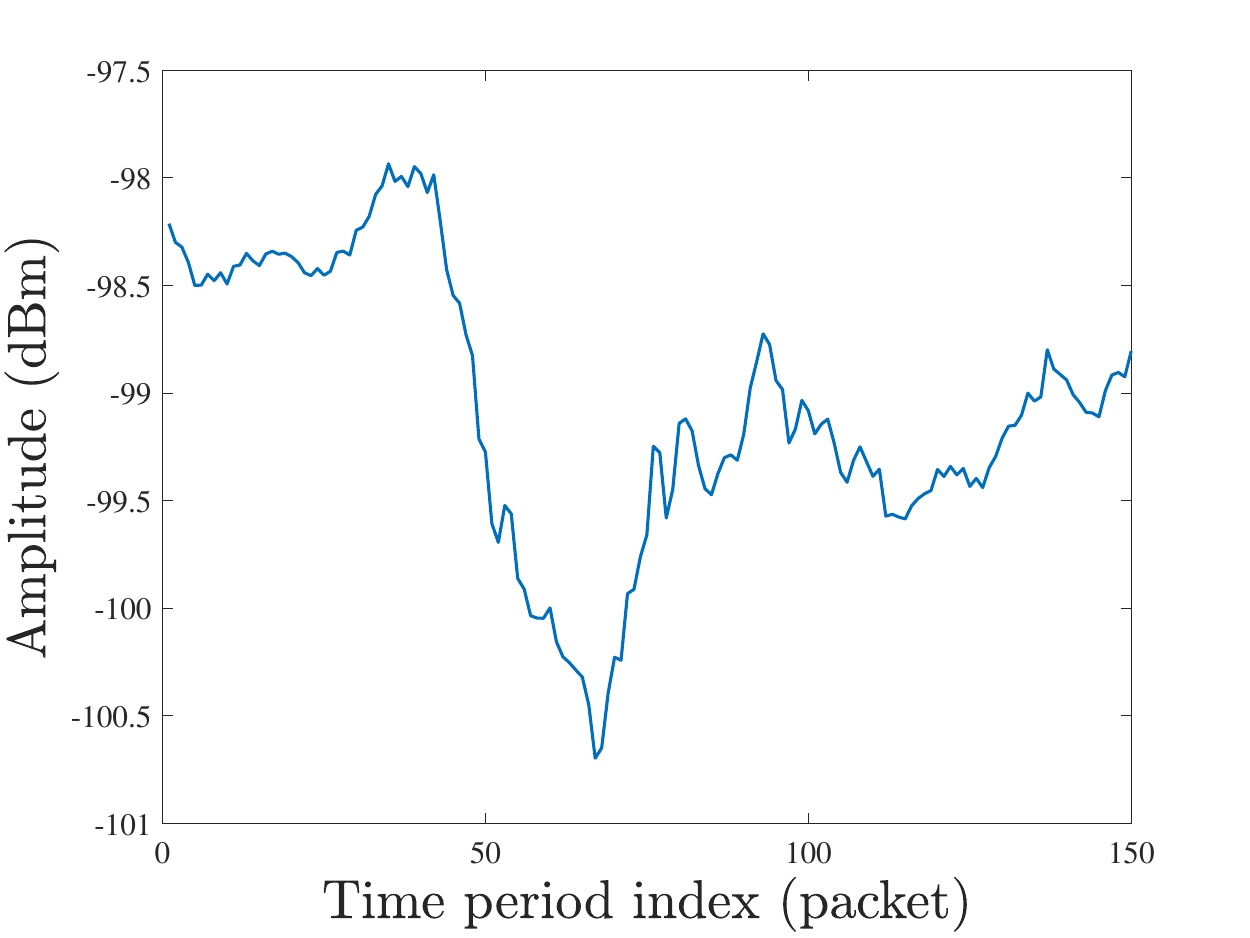}
	\end{minipage}
	\begin{minipage}{0.22\linewidth}
		\centering
		\includegraphics[width=1\linewidth]{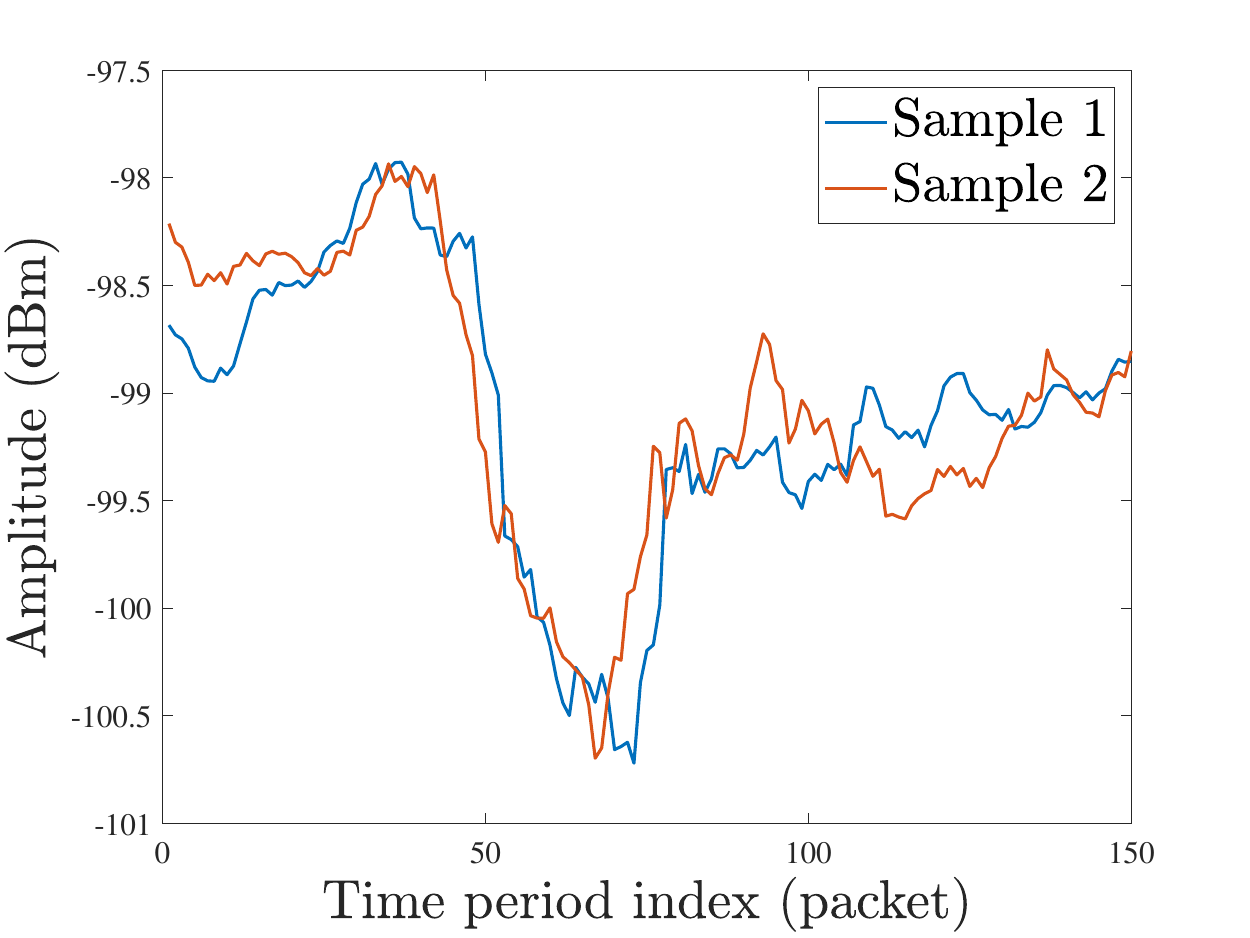}
	\end{minipage}
	\begin{minipage}{0.22\linewidth}
		\centering
		\includegraphics[width=1\linewidth]{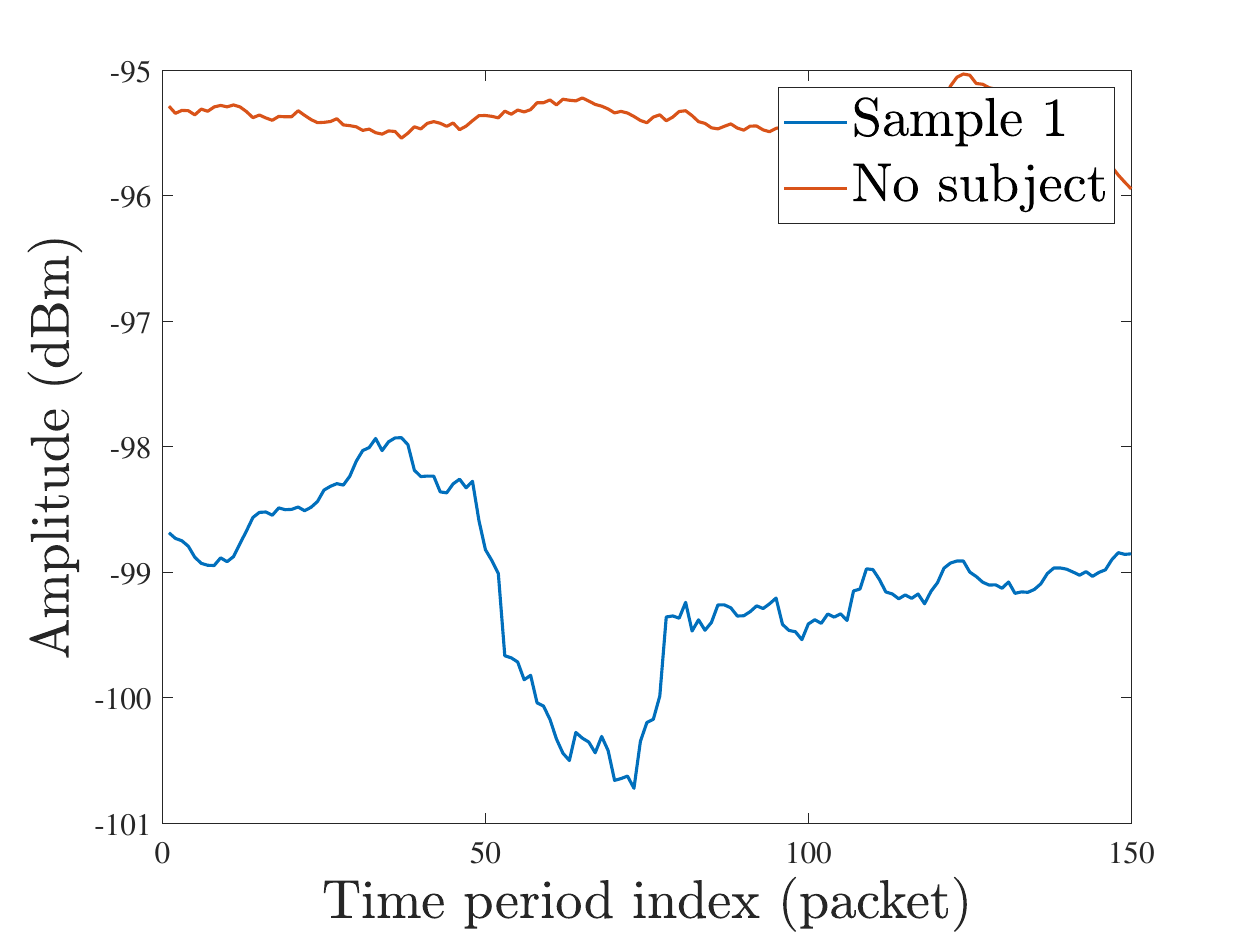}
	\end{minipage}
    \caption{CSI data of different subjects and vacant situation.} 
    \label{figure:walk_compare}
\end{figure*}

In order to analyze the effect of gait and different persons on CSI, we visualize
one of the CSI subcarriers (the 5200th out of 8192) for analysis. We
select some CSI samples of two persons as shown in Fig.~\ref{figure:walk_compare}, where the $y-axis$ is the amplitude of the 5200th subcarrier and the $x-axis$ is the index of the time packet. By comparing the figures with person and no person, it is observed that the presence of person leads to obvious CSI variations.
By comparing the figures of different persons, it has been observed that the CSI patterns of the same person are consistent and the CSI patterns of the different persons are inconsistent, demonstrating that the CSI is capable of reflecting the ID of a distinct person. This consistency in CSI patterns provides a reliable basis for developing a gait recognition system that leverages wireless signals. 
\subsection{Feature Extraction and Classification}
The architecture and parameters of the RCNN are presented in Table~\ref{network:RCNN}.
The RCNN is composed of several key components, including convolution blocks, batch normalization layers, ReLU layers, max-pooling layers, a flatten layer, and a fully connected (FC) layer. 

The amplitude information from CSI is input to the network and passed through a convolution block with a $3\times3$ kernel and $2\times2$ stride. 
A batch normalization layer and a ReLU layer are then applied to improve convergence.
The batch normalization layer addresses the issue of
gradient vanishing and improves performance\cite{pmlr-v37-ioffe15}. 
The ReLU layer adds nonlinearity to increase model capacity.
To extract deeper features, we design two residual blocks\cite{He_2016_CVPR} and three convolution blocks with 16, 8, and 8 channels. Except for the last convolutional block, each convolution block is followed by a maximum pool layer \cite{5539963} with a kernel of $2\times2$ and a stride of $2\times2$. This helps to reduce the number of parameters in the network and improve the robustness and generalization capability of our model. Finally, one FC layer and softmax function convert extracted features into a $K$-dimensional vector, where $K$ represents the number of individuals.
\begin{table}[!ht]
    \renewcommand{\arraystretch}{1.5}
    \caption{Network Architecture of RCNN(
    \textit{Ch} Represents the Channel)}
    \label{network:RCNN}
    \centering
    \begin{tabular}{lcc}
    \hline
        \textbf{Input} & \multicolumn{2}{c}{$x_{c}\in R^{1\times150\times8192}$} \\ \hline \midrule 
        Block Name&Output Size&Parameters\\ \hline
        Conv 1&$8\times74\times4095$&$3\times3,2,Ch 8$\\ \hline
        Res-block 1&$8\times37\times2048$&$\begin{bmatrix}3\times3,2,8\\3\times3,2,8\end{bmatrix}$\\ \hline 
        Res-block 2&$8\times37\times2048$&$\begin{bmatrix}3\times3,2,8\\3\times3,2,8\end{bmatrix}$\\ \hline 
        Pooling-layer 1&$8\times18\times1024$&$2,2$\\ \hline
        Conv 2&$16\times8\times511$&$3\times3,2,Ch 16$\\ \hline
        Pooling-layer 2&$16\times4\times255$&$2,2$\\ \hline
        Conv 3&$8\times2\times128$&$3\times3,2,Ch 8$\\ \hline
        Pooling-layer 3&$8\times1\times64$&$2,2$\\ \hline
        Conv 4&$8\times1\times32$&$3\times3,2,Ch 8$\\ \hline
        FC&$K$&Linear,$K$\\ \hline
        \textbf{Output}& \multicolumn{2}{c}{$\hat{y}\in R^K$} \\ \hline
    \end{tabular}
\end{table}
\section{Evaluation}

\subsection{Experiment Setup}
\begin{figure}[!ht]
\centering
\subfigure[Meeting Room]
{
 	\begin{minipage}[b]{.28\linewidth}
        \centering
        \includegraphics[width=1\linewidth]{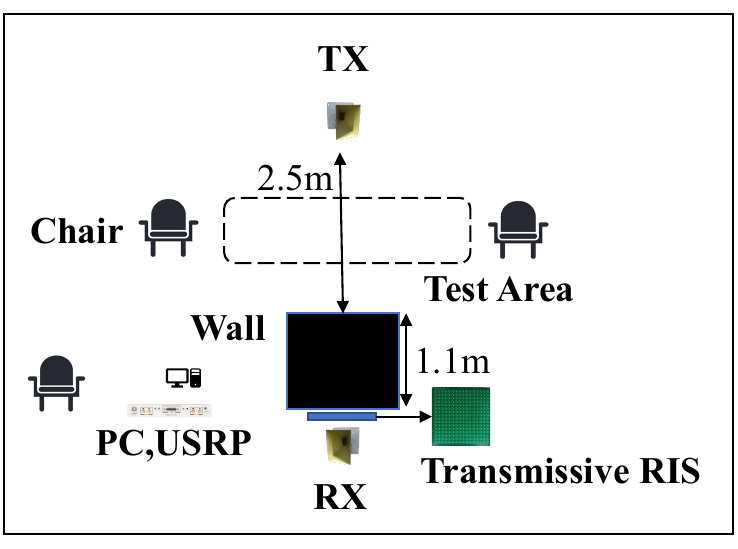}
    \end{minipage}
}
\subfigure[Laboratory]
{
 	\begin{minipage}[b]{.28\linewidth}
        \centering
        \includegraphics[width=1\linewidth]{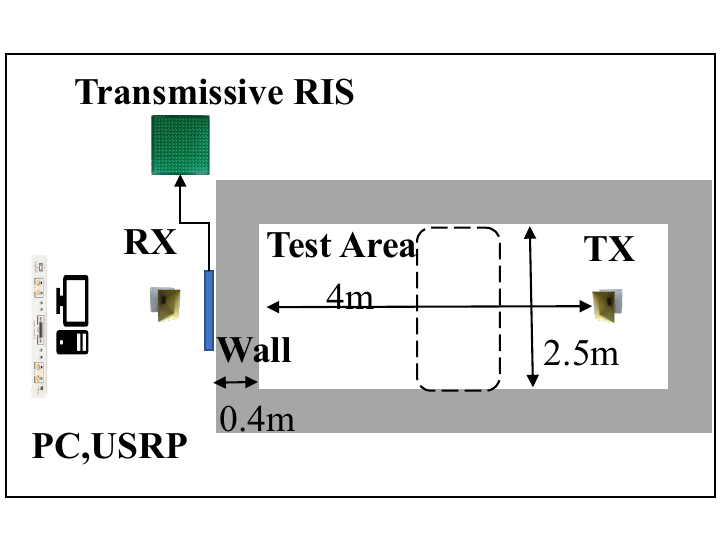}
    \end{minipage}
}
\subfigure[Office]
{
 	\begin{minipage}[b]{.28\linewidth}
        \centering
        \includegraphics[width=1\linewidth]{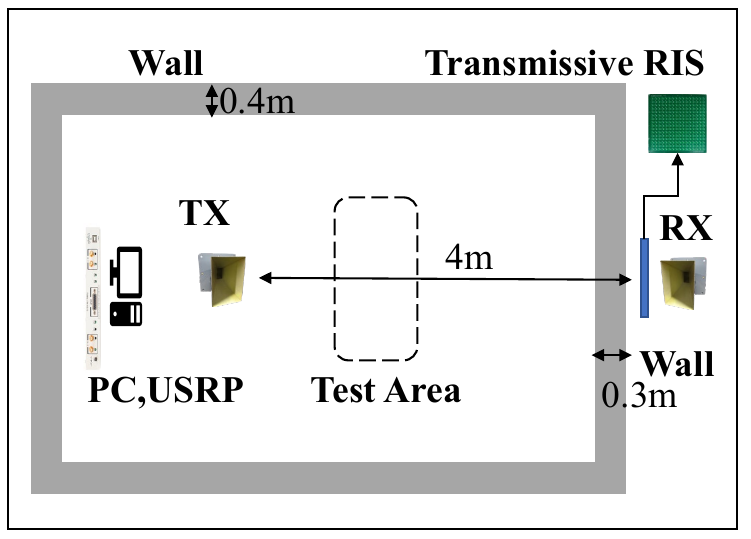}
    \end{minipage}
}
\caption{Different data acquisition scenarios.}
\label{figure:different_scenarios}
\end{figure}

To evaluate the performance of TRGR, we collect data on three scenarios: a meeting room, laboratory, and office, depicted in Fig.~\ref{figure:different_scenarios}, with 5, 6, and 4 volunteers participating in each scenario, respectively.

In comparative experiments, empirical data underscores the superior performance of TRGR against traditional human sensing approaches. The methodology involves recording a person walking from one side to another as a distinct sample.
For the meeting room, laboratory, and office scenarios, 100, 50, and 50 samples are respectively collected per person. The gait of each person is captured for 3 seconds. The $2/3$ of samples are allocated to the training set, while the remaining $1/3$ is assigned to the testing set. The system records data at a rate of 50 packets per second at the receiver, resulting in 150 packets of CSI data over a 3-second sensing period. 
The data consists of 8192 subcarriers, thereby,  the size of each CSI data is $1\times150\times8192$.

The learning scheme of TRGR is implemented by PyTorch, and the model is trained on one NVIDIA GeForce GTX 3090Ti. The Adam optimizer is leveraged for better convergence. 
The batch size is set to 8 with a learning rate of $10^{-3}$ and a total of 20 epochs. The performance of the proposed TRGR is evaluated by the accuracy, recall, precision, and F1 score.
\subsection{Performance Evaluation}
\subsubsection{Impact of Transmissive RIS} %
\begin{figure}[!ht]
\centering
\subfigure[Without transmissive RIS]
{
    \begin{minipage}[b]{0.46\linewidth}
        \centering
        \includegraphics[width=1\linewidth]{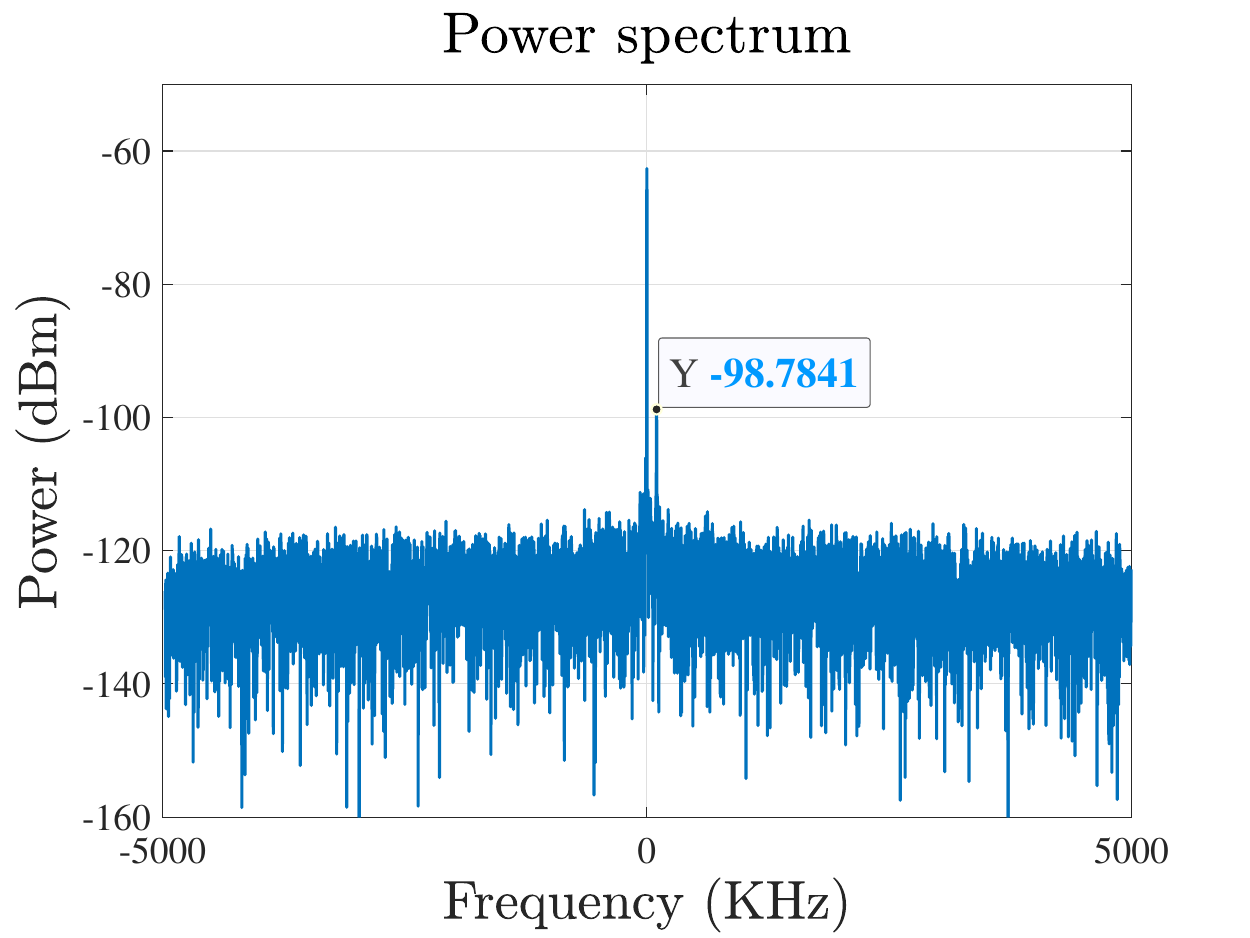}
    \end{minipage}
}
\subfigure[With transmissive RIS]
{
    \begin{minipage}[b]{0.46\linewidth}
        \centering
        \includegraphics[width=1\linewidth]{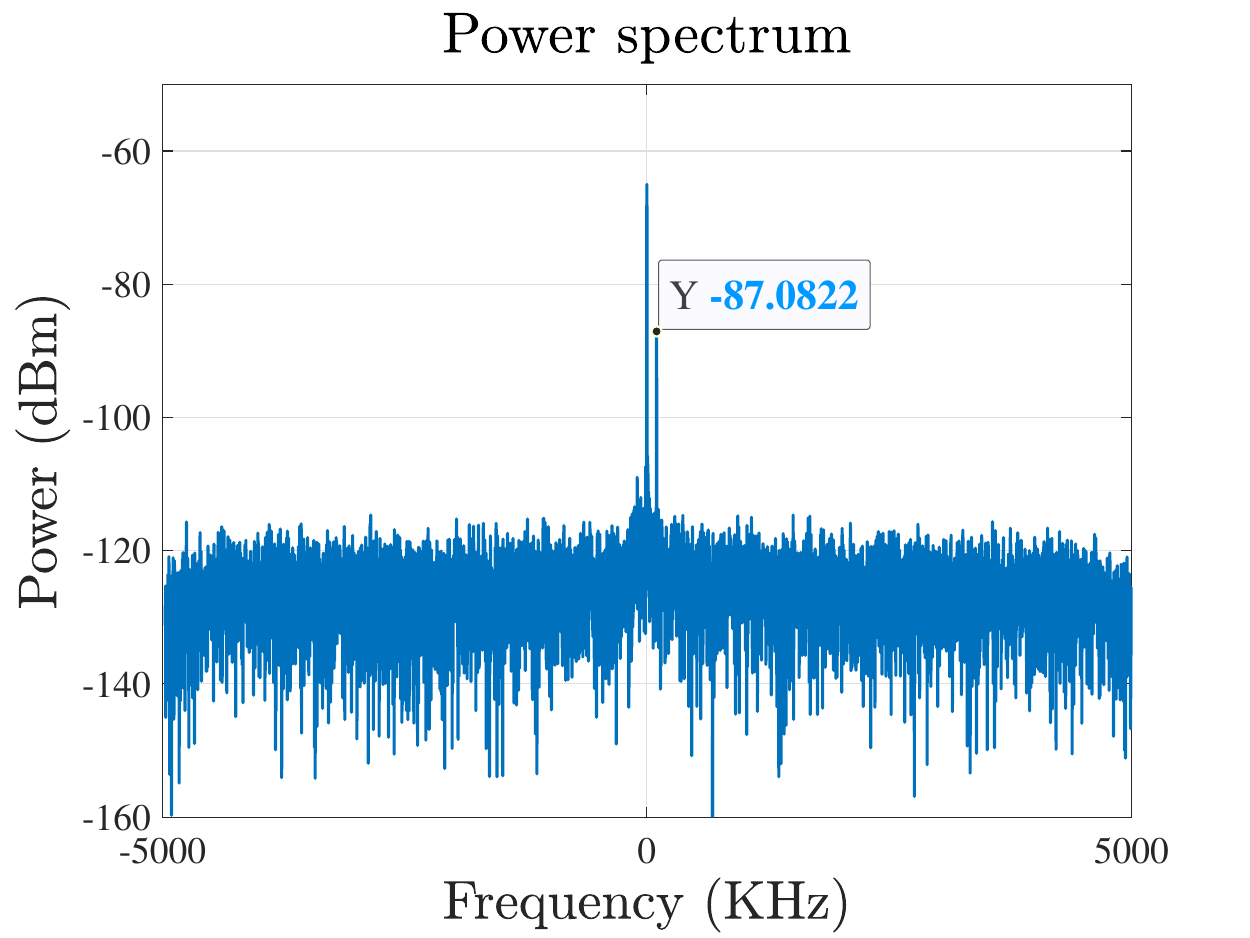}
    \end{minipage}
}
\caption{Effects of without RIS and with transmissive RIS on the narrowband signal strength}
\label{figure:narrowband_enhance}
\end{figure}

To evaluate the impact of transmissive RIS on the system, we conduct a comparative analysis with and without the incorporation of transmissive RIS, specifically focusing on narrowband signal strength and system accuracy. 
The power spectrum of the narrowband signal under both conditions is illustrated in Fig.~\ref{figure:narrowband_enhance}. It is observed that the signal strength is boosted by approximately 10dB when utilizing transmissive RIS compared to the scenario without it, highlighting the effectiveness of our transmissive RIS in improving SNR. 

Additionally, to fully elucidate the role of transmissive RIS, we conducted experiments in a laboratory setting that exhibited low SNR in the absence of transmissive RIS.
The comparative results are summarized in Table~\ref{tab:ris_res}. With the enhancement of transmissive RIS, the accuracy is improved significantly by 11.46\%, underscoring the significant role of transmissive RIS in enhancing the performance of the system.

\begin{table}[!ht]
    \caption{Impact of Transmissive RIS}
    \label{tab:ris_res}
    \centering
    \resizebox{1\linewidth}{!}{
    \begin{tabular}{cccccc}
    \hline
        \makecell[c]{With or without\\transmissive RIS}&Accuracy$(\%)$& Recall(\%) & Precision(\%) & F1 score(\%)\\ \hline
        \ding{53}&87.50 & 87.50 & 89.67 & 87.34\\   \textbf{\ding{51}}&\textbf{98.96}&\textbf{98.96}&\textbf{99.02}&\textbf{98.96}\\\hline
    \end{tabular}}
\end{table}
\subsubsection{Comparisons of Different Models}
To compare the RCNN model with other models, we conducted an experiment in a meeting room scenario with transmissive RIS. We compare our method with novel WiFi-based human sensing methods, including CNN-GRU \cite{Shalaby2022}, CNN-GRU-CNN \cite{Shalaby2022}, LSTM-CNN \cite{Shang_2021} and GaitFi \cite{9887951}, as shown in Table~\ref{tab:compare_model}.

The findings underscore the exceptional performance of RCNN, achieving the highest accuracy of 96.25\%. This result not only illustrates the superior capability of RCNN in accurately identifying individuals based on gait but also underscores the significant advancement it represents over existing models within WiFi-based human sensing technologies.
\begin{table}[!ht]
    \centering
    \caption{Comparisons of Different Models}
    \resizebox{1\linewidth}{!}{
    \begin{tabular}{ccccc}
    \hline
        Model & Accuracy(\%) & Recall(\%) & Precision(\%) & F1 score(\%) \\ \hline
        CNN-GRU &62.50&62.50&68.29&61.08\\ 
        CNN-GRU-CNN &61.88&61.88&67.05&58.60\\ 
        LSTM-CNN &69.38&69.38&74.02&69.49\\ 
        GaitFi  &80.00&80.00&70.00&73.33\\ 
        \textbf{RCNN}&\textbf{96.25}&\textbf{96.25}&\textbf{96.84}&\textbf{96.22}\\ \hline
    \end{tabular}
    \label{tab:compare_model}
    }
\end{table}
\subsubsection{Model Adaptivity in Different Scenarios} 
To evaluate the model adaptivity to different multipath effects, we establish three scenarios as depicted in Fig.~\ref{figure:different_scenarios}.
RCNN is trained and tested by CSI data collected in each scenario separately.
We vary the scenarios by building area, equipment layout, and room functionality to ensure significant differences in multipath effects on CSI. 
\begin{table}[!ht]
  \centering
  \caption{Model Performance in Different Scenarios}
  \label{tab:different_scenarios}
  \resizebox{1\linewidth}{!}{
  \begin{tabular}{ccccc}
    \hline
        Scenario & Accuracy(\%) & Recall(\%) & Precision(\%) & F1 score(\%) \\ \hline
        Meeting Room&96.25&96.25&96.84&96.22 \\ 
        Laboratory&98.96&98.96&99.02&98.96\\ 
        Office&98.44&98.44&98.53&98.44\\ \hline
    \end{tabular}}
\end{table}

As shown in Table~\ref{tab:different_scenarios}, TRGR achieves an accuracy of 96.25\% in the meeting room scenario, demonstrating the superior performance of TRGR. For a closed and surrounded-by-walls laboratory scenario, our proposed RCNN ensures promising performance with an accuracy of 98.96\%. Furthermore, for an office scenario, RCNN ensures promising performance with an accuracy of 98.44\% as well. 
These findings highlight the adaptability and potential of the RCNN model in diverse environments, characterized by varying multipath effects.
\section{Conclusion}
In this article, we present TRGR, a novel transmissive RIS-aided through-wall gait recognition system, which recognizes human identities through walls using only the CSI magnitude measurements from transceivers.
Specifically, by leveraging transmissive RIS alongside a configuration alternating optimization algorithm, TRGR enhances wall penetration and signal quality. Furthermore, RCNN is proposed as the backbone network to learn robust human information, which can accurately identify individuals. 
Experimental results confirm the efficacy of transmissive RIS, highlighting the significant potential of transmissive RIS in enhancing RF-based gait recognition systems.
Extensive experiment results show that TRGR achieves an average accuracy of 97.88\% in identifying persons when signals traverse concrete walls, demonstrating the effectiveness and robustness of TRGR. This study not only showcases the feasibility and effectiveness of TRGR in overcoming physical obstructions for gait recognition but also underscores its potential application in intelligent IoT environments and smart spaces.
\section*{Acknowledgment}
This work was supported in part by the Nation Natural Science Foundation of China under Grant No.12141107, in part by the Guangxi Science and Technology Project (AB21196034), and in part by the Interdisciplinary Research Program of HUST, 2023JCYJ012.
\bibliographystyle{IEEEtran}
\bibliography{main}

\begin{thebibliography}{10}
\providecommand{\url}[1]{#1}
\csname url@samestyle\endcsname
\providecommand{\newblock}{\relax}
\providecommand{\bibinfo}[2]{#2}
\providecommand{\BIBentrySTDinterwordspacing}{\spaceskip=0pt\relax}
\providecommand{\BIBentryALTinterwordstretchfactor}{4}
\providecommand{\BIBentryALTinterwordspacing}{\spaceskip=\fontdimen2\font plus
\BIBentryALTinterwordstretchfactor\fontdimen3\font minus \fontdimen4\font\relax}
\providecommand{\BIBforeignlanguage}[2]{{%
\expandafter\ifx\csname l@#1\endcsname\relax
\typeout{** WARNING: IEEEtran.bst: No hyphenation pattern has been}%
\typeout{** loaded for the language `#1'. Using the pattern for}%
\typeout{** the default language instead.}%
\else
\language=\csname l@#1\endcsname
\fi
#2}}
\providecommand{\BIBdecl}{\relax}
\BIBdecl

\bibitem{ali2016overview}
M.~M. Ali, V.~H. Mahale, P.~Yannawar, and A.~T. Gaikwad, ``Overview of fingerprint recognition system,'' in \emph{2016 International Conference on Electrical, Electronics, and Optimization Techniques (ICEEOT)}, 2016, pp. 1334--1338.

\bibitem{Mehmood2010}
A.~Mehmood, J.~M. Sabatier, M.~Bradley, and A.~Ekimov, ``Extraction of the velocity of walking human’s body segments using ultrasonic {Doppler},'' \emph{The Journal of the Acoustical Society of America}, vol. 128, pp. EL316--EL322, 11 2010.

\bibitem{Wang2014}
F.~Wang, M.~Skubic, M.~Rantz, and P.~E. Cuddihy, ``{Quantitative Gait Measurement With Pulse-Doppler Radar for Passive In-Home Gait Assessment},'' \emph{IEEE Transactions on Biomedical Engineering}, vol.~61, no.~9, pp. 2434--2443, 2014.

\bibitem{5766971}
T.~Yardibi, P.~Cuddihy, S.~Genc, C.~Bufi, M.~Skubic, M.~Rantz, L.~Liu, and C.~Phillips, ``{Gait characterization via pulse-Doppler radar},'' in \emph{2011 IEEE International Conference on Pervasive Computing and Communications Workshops (PERCOM Workshops)}, 2011, pp. 662--667.

\bibitem{wang2016gait}
W.~Wang, A.~X. Liu, and M.~Shahzad, ``{Gait recognition using wifi signals},'' in \emph{Proceedings of the 2016 ACM International Joint Conference on Pervasive and Ubiquitous Computing}, 2016, pp. 363--373.

\bibitem{7460727}
Y.~Zeng, P.~H. Pathak, and P.~Mohapatra, ``{WiWho: WiFi-Based Person Identification in Smart Spaces},'' in \emph{2016 15th ACM/IEEE International Conference on Information Processing in Sensor Networks (IPSN)}, 2016, pp. 1--12.

\bibitem{7536315}
J.~Zhang, B.~Wei, W.~Hu, and S.~S. Kanhere, ``{WiFi-ID: Human Identification Using WiFi Signal},'' in \emph{2016 International Conference on Distributed Computing in Sensor Systems (DCOSS)}, 2016, pp. 75--82.

\bibitem{tang2023transmissive}
J.~Tang, M.~Cui, S.~Xu, L.~Dai, F.~Yang, and M.~Li, ``{Transmissive RIS for B5G Communications: Design, Prototyping, and Experimental Demonstrations},'' \emph{IEEE Transactions on Communications}, vol.~71, no.~11, pp. 6605--6615, 2023.

\bibitem{yang2018device}
J.~Yang, H.~Zou, H.~Jiang, and L.~Xie, ``{Device-Free Occupant Activity Sensing Using WiFi-Enabled IoT Devices for Smart Homes},'' \emph{IEEE Internet of Things Journal}, vol.~5, no.~5, pp. 3991--4002, 2018.

\bibitem{pei2021ris}
X.~Pei, H.~Yin, L.~Tan, L.~Cao, Z.~Li, K.~Wang, K.~Zhang, and E.~Bj{\"o}rnson, ``Ris-aided wireless communications: Prototyping, adaptive beamforming, and indoor/outdoor field trials,'' \emph{IEEE Transactions on Communications}, vol.~69, no.~12, pp. 8627--8640, 2021.

\bibitem{ettus2015universal}
M.~Ettus and M.~Braun, ``The {Universal} {Software} {Radio} {Peripheral} ({USRP}) {Family} of {Low-Cost} {SDRs},'' \emph{Opportunistic spectrum sharing and white space access: The practical reality}, pp. 3--23, 2015.

\bibitem{Kodosky2020}
J.~Kodosky, ``{LabVIEW},'' \emph{Proceedings of the ACM on Programming Languages}, vol.~4, 6 2020.

\bibitem{7581108}
E.~Isufi, A.~Loukas, A.~Simonetto, and G.~Leus, ``{Autoregressive Moving Average Graph Filtering},'' \emph{IEEE Transactions on Signal Processing}, vol.~65, no.~2, pp. 274--288, 2017.

\bibitem{pmlr-v37-ioffe15}
S.~Ioffe and C.~Szegedy, ``{Batch Normalization: Accelerating Deep Network Training by Reducing Internal Covariate Shift},'' in \emph{Proceedings of the 32nd International Conference on Machine Learning}, ser. Proceedings of Machine Learning Research, F.~Bach and D.~Blei, Eds., vol.~37.\hskip 1em plus 0.5em minus 0.4em\relax PMLR, 07--09 Jul 2015, pp. 448--456.

\bibitem{He_2016_CVPR}
K.~He, X.~Zhang, S.~Ren, and J.~Sun, ``{Deep Residual Learning for Image Recognition},'' in \emph{Proceedings of the IEEE Conference on Computer Vision and Pattern Recognition (CVPR)}, June 2016.

\bibitem{5539963}
Y.-L. Boureau, F.~Bach, Y.~LeCun, and J.~Ponce, ``{Learning mid-level features for recognition},'' in \emph{2010 IEEE Computer Society Conference on Computer Vision and Pattern Recognition}, 2010, pp. 2559--2566.

\bibitem{Shalaby2022}
E.~Shalaby, N.~ElShennawy, and A.~Sarhan, ``{Utilizing deep learning models in CSI-based human activity recognition},'' \emph{Neural Computing and Applications}, vol.~34, pp. 5993--6010, 4 2022.

\bibitem{Shang_2021}
S.~Shang, Q.~Luo, J.~Zhao, R.~Xue, W.~Sun, and N.~Bao, ``{LSTM-CNN network for human activity recognition using WiFi CSI data},'' \emph{Journal of Physics: Conference Series}, vol. 1883, no.~1, p. 012139, apr 2021.

\bibitem{9887951}
L.~Deng, J.~Yang, S.~Yuan, H.~Zou, C.~X. Lu, and L.~Xie, ``{GaitFi: Robust Device-Free Human Identification via WiFi and Vision Multimodal Learning},'' \emph{IEEE Internet of Things Journal}, vol.~10, no.~1, pp. 625--636, 2023.

\end{thebibliography}
\end{document}